\documentclass[10pt,onecolumn]{article}

\usepackage{cvpr}
\usepackage{times}
\usepackage{graphicx}
\usepackage{amsmath}
\usepackage{amssymb}

\usepackage[pagebackref=true,breaklinks=true,letterpaper=true,colorlinks,bookmarks=false]{hyperref}

\cvprfinalcopy 

\def\cvprPaperID{2546} 

\begin{document}

\title{Modeling Information Flow Through Deep Neural Networks}

\author{Ahmad Chaddad
\and
Marco Pedersoli
\and
Eric Granger
\and
Christian Desrossiers
\and
Matthew Toews
\\
\'Ecole de T\'echnologie Sup\'erieure\\
Montr\'eal, Qu\'ebec, Canada\\
{\tt\small matt.toews@gmail.com}
}


\maketitle

\begin{abstract}
This paper proposes a principled information theoretic analysis of classification for deep neural network structures, e.g. convolutional neural networks (CNN). The output of convolutional filters is modeled as a random variable $Y$ conditioned on the object class $C$ and network filter bank $F$. The conditional entropy (CENT) $H(Y\, | \,C,F)$ is shown in theory and experiments to be a highly compact and class-informative code, that can be computed from the filter outputs throughout an existing CNN and used to obtain higher classification results than the original CNN itself. Experiments  demonstrate the effectiveness of CENT feature analysis in two separate CNN classification contexts. 1) In the classification of neurodegeneration due to Alzheimer's disease (AD) and natural aging from 3D magnetic resonance image (MRI) volumes, 3 CENT features result in an AUC=94.6\% for whole-brain AD classification, the highest reported accuracy on the public OASIS dataset used and 12\% higher than the softmax output of the original CNN trained for the task. 2) In the context of visual object classification from 2D photographs, transfer learning based on a small set of CENT features identified throughout an existing CNN leads to AUC values comparable to the 1000-feature softmax output of the original network when classifying previously unseen object categories. The general information theoretical analysis explains various recent CNN design successes, e.g. densely connected CNN architectures, and provides insights for future research directions in deep learning.
\end{abstract}

\section{Introduction}

The seminal work of Kreshevsky and Hinton~\cite{krizhevsky2012imagenet} showed that deep convolutional neural networks (CNN) ~\cite{lecun1989backpropagation,lecun1998gradient} powered by highly parallelized graphics processing units (GPU) could be used to achieve unprecedented performance in image classification. This resulted in a paradigm shift in computer vision and machine learning that has continuously improved upon the CNN technology.

Many improvements have involved redesigning or modifying the original CNN framework, for example ReLU activation, max pooling, dropout, variations in loss functions, transfer learning \cite{oquab2014learning}, network topology or structure, e.g. residual \cite{he2016identity} or densely connected networks \cite{huang2016densely}. Reduction of memory and computation time has also been an important focus, obtaining similar classification rates at reduced levels of memory and algorithmic complexity~\cite{rastegari2016xnor,song2015learning}. 
While these and other developments have had a marked improvement on CNN classification performance, many are the result of heuristic or ad-hoc design insights that are difficult to understand within a single unified theoretical framework.

This paper provides a principled analysis of the CNN within the framework of information theory~\cite{shannon2001mathematical} that, to our knowledge, is novel in the context of recent CNN research. We propose quantifying the flow of information through network filters in terms of the conditional entropy $H(Y\, | \,C,F)$ of neural activation $Y$, given object class $C$ and filtering operation $F$. A set of conditional entropy (CENT) features derived from a trained CNN is shown in theory and experiments to be a highly informative and compact code for classification, improving upon the output of the original CNN from which the features were originally derived. Very recent research has begun investigating entropy through network nodes~\cite{shwartz2017opening,kuo2017data}. With respect to this research, we provide a theoretical proof that conditional information $H(Y\, | \,C,F)$ is {\em necessarily} a highly informative code for discriminate filter set $F$, and we show how very small sets of CENT features can be identified and used as class-informative codes.

Experiments demonstrate CENT analysis in the context of CNN classification of natural aging and Alzheimer's disease from volumetric magnetic resonance images (MRI) of the human brain and of visual object categories in 2D photographs. A basic 4-layer CNN architecture produces state-of-the-art performance on the publicly available OASIS neuroimage data ~\cite{marcus2007open} from 3 CENT features. For an existing CNN trained to classify a large set of visual object categories, several CENT features computed an existing CNN can be used to classifying new objects not used in CNN training with AUC values comparable to the 1000-feature softmax output of the original CNN. Aside from CENT features, the general information theoretical framework we develop is useful in understanding the success of recent developments in CNN technology, including the high performance of densely connected networks\cite{huang2016densely} and provides new insights and directions for future research.

\section{Previous Work}

Information flows through a neural network, transforming input data into recognizable symbols (i.e., object class labels) at the output. Along the path are neurons, units that integrate and filter information before passing it on to other neurons. Our analysis links basic yet powerful principles in information theory and CNN technology, to derive a set of highly informative features for CNN-based image classification. In order to best assist the understanding of the reader, we forgo the daunting task of a comprehensive literature review of all important developments in deep CNN technology, and rather restrict our references to directly related theoretical work and major related developments.

\textbf{Information theory} was first developed in the work of Shannon~\cite{shannon2001mathematical} and now serves as the basis of all modern digital communication systems ~\cite{cover2012elements}. As deep CNNs involve information flow through networks, it is natural to analyze them in terms of information theory \cite{murata1994network,bell1997independent}. Here, we define information theoretical concepts used in our analysis that can generally be found in a suitable textbook on the subject~\cite{cover2012elements}. Information theory is rooted in the notion of entropy, which quantifies the uncertainty of a random variable $Y=\{y_1, \dots, y_i, \dots, y_N\}$ defined over a set of $N$ discrete events $y_i$ which occur with probability $p(Y=y_i)$, or $p(y_i)$. The Shannon entropy $H(Y)$ is defined as
\begin{align}
	H(Y) \ = \ - \sum_{i=1}^N p(y_i) \, \log p(y_i),
\end{align}
and ranges from $[0,\log N]$ for maximally informative and uninformative distributions $p(Y)$, respectively. For a binary random variable, the entropy is proportional to the expected number of bits required to transmit 1 symbol of information. 
Entropy has been widely used in computer vision, e.g., in classifying image textures~\cite{haralick1973textural}, or salient feature selection~\cite{kadir2001saliency,toews2003entropy}.

Given a second random variable, e.g., object class $C=\{c_1, \dots, c_j, \dots c_M\}$, the conditional entropy is defined as
\begin{align}
	H(Y\, | \,C) \ = \ \sum_{j=1}^M p(c_j) \, H(Y\, | \,c_j),
\end{align}
where $H(Y\, | \,c_j)$ is the entropy of $Y$ conditioned on a fixed class $c_j$, and $H(Y\, | \,C)$ thus represents the expected conditional entropy across all classes. An important consequence is that entropy is reduced by conditioning $H(Y\, | \,C) \le H(Y)$ except in the case where $Y$ and $C$ are statistically independent, in which case the equality holds. The difference between the entropy and the conditional entropy provides the amount of information shared by $Y$ and $C$, and is known as the mutual information (MI) $I(Y,C)$:
\begin{align}
	I(Y,C) \, = \, H(Y)-H(Y\, | \,C) \, = \, H(C)-H(C\, | \,Y).
\end{align}
MI provides an upper bound as to the capacity of a noisy communication channel $Y \rightarrow C$~\cite{cover2012elements}. Conditional entropy and MI are widely used to measure image-to-image similarity in computer vision~\cite{wells1996multi} and as a feature selection strategy in machine learning~\cite{brown2012conditional}. For example, in decision tree learning, MI is known as the Information Gain~\cite{duda2012pattern} and used to identify optimal data splits in training. In contrast, our analysis involves computing conditional entropy from a suitably trained, pre-existing CNN.

A final definition is the data processing inequality, which states that if $X \rightarrow Y \rightarrow C$ is a Markov chain, then $I(Y,C) \le I(X,C)$, i.e. the information shared between endpoints of a network $I(X,C)$ is generally greater than that of any intermediate node and the endpoint $I(Y,C)$~\cite{cover2012elements}. When applied to CNNs, this result explains the success of densely connected networks~\cite{huang2016densely}, which are able to furnish a greater amount of information to the final classification layer than standard sequential processing, thanks to additional links between all layers and the network output. We apply this insight by performing classification based on highly informative CENT features, at a tiny fraction of the data and computation required for dense CNN modeling. 


{\bf Deep neural networks} arose from multi-layered perceptron networks, where weight parameters were trained via the backpropagation algorithm~\cite{rosenblatt1961principles} used ubiquitously today. Deep convolutional neural networks (CNN) were introduced as an efficient multi-layer perceptron approach for image data, due to smaller, translation invariant image filters and came into prominence in the context of text and document analysis~\cite{lecun1989backpropagation,lecun1998gradient}.
A major development was the use of parallel graphics processing units (GPU), which allowed training on large-scale data sets~\cite{krizhevsky2012imagenet}. While fundamental aspects of CNN technology remain layers of image filters trained via backpropagation, various algorithmic improvements have been introduced such as dropout~\cite{sriva2014dropout}, batch normalization~\cite{sergey2015batch}, improved pooling~\cite{graham2014fractional}, different activation units~\cite{clevert2015fast} and better topologies~\cite{huang2016densely}.

CNN training often makes use of the so-called cross-entropy loss function, also known as the log loss function. The log loss is the derivative of the multivariate softmax function typically found at the CNN output, and is used as an error signal to update CNN weights during backpropagation training. In contrast, our analysis considers conditional entropy computed across all layers and/or filters of an existing CNN, and applies generally to CNNs trained with any suitable output loss function, including the cross-entropy, squared or absolute loss, etc, the only requirement being that the resulting filters be informative regarding the object classes of interest. Our analysis predicts that the conditional entropy (CENT) can serve as a highly informative feature for classification.


\section{Conditional Entropy (CENT) Analysis}

Let $Y$ a scalar random variable representing the output of neurons in a CNN, e.g., responses in a feature map. The distribution $p(Y)$ of informative convolution filters can be approximated by a heavy-tailed distribution such as a Laplacian due to the correlation structure of natural images~\cite{simoncelli2001natural,simon2007scene}. While $Y$ is a continuous random variable, we approximate it as a discrete random variable for the purpose of computing Shannon entropy. 

Let $C = \{c_j\}$ be a discrete random variable over object classes. Classification typically seeks to maximize the Bayesian posterior distribution $p(C\, | \,Y,F) \propto p(Y\, | \,C,F)\, p(C\, | \,F)$ over possible classes $C$ given filtered image data $Y$ and a set of filtering operations $F$, i.e., maximum a-posteriori (MAP) estimation. In the posterior, $p(Y\, | \,C,F)$ represents the conditional likelihood of class $C$ associated with data $Y$ and filter set $F$ and $p(C\, | \,F)$ is a conditional prior over classes $C$ given feature set $F$. For the purpose of our analysis, we focus on the data term $p(Y\, | \,C,F)$, which models the link between data $Y$ and class $C$, and assume a uniform distribution $p(C\, | \,F)$.

In the case of uniform $p(C\, | \,F)$, MAP estimation is equivalent to maximum likelihood (ML) estimation, and classification seeks to identify a distribution $p(Y\, | \,C,F)$ of $Y$ conditioned on class $C$ and filter set $F$ such probabilistic uncertainty is minimized, i.e., such that conditional entropy $H(Y\, | \,C,F)$ is minimal. In the case of $H(Y\, | \,C,F) = 0$, perfect classification accuracy may be achieved via ML estimation, i.e. $C^\mathrm{ML} = \textrm{argmax}_C \, p(Y\, | \,C, F)$. 

\begin{figure}[t]
\begin{center}
\includegraphics[width=0.6\linewidth]{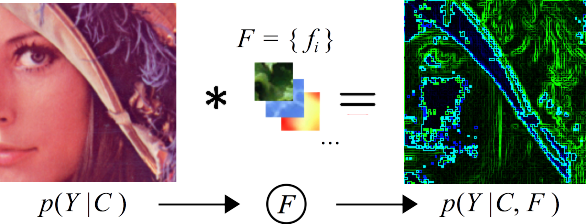}
\end{center}
   \caption{The operation of a set of potentially non-linear filters $F$ is modeled as a conditional probability $p(Y\, | \,C,F)$ over incoming data $Y$ arising from an object of class $C$ .}
\label{fig:filtering}
\end{figure}

Let $F=\{f_i\}$ be a discrete random variable over a set of filtering operators $f_i$ applied at neurons. As shown in Figure~\ref{fig:filtering}, the result of a filtering operation may be expressed as a conditional distribution $p(Y\, | \,C,F)$ over neural output $Y$ conditioned on class $C$ and filter $F$. Modeling filtering via probabilistic conditioning has a rich history in signal processing, e.g., the linear Kalman filter~\cite{kalman1960new}, and generally applies to both arbitrary linear and non-linear filtering operations, e.g., convolution and/or ReLU in CNNs. In the context of information theory and classification, the primary importance of filtering is to reduce the entropy $H(Y\, | \,C,F)$ by conditioning on a judiciously designed filter set $F$. Specifically 
\begin{align}
  H(Y\, | \,C) & \ \geq \ H(Y\, | \,C,F) \\ \notag 
     & \ = \ \sum_{i,j} p(c_j,f_i) \, H(Y\, | \,c_j,f_i),
\end{align}
where $H(Y\, | \,C,F)$ is the conditional entropy of $Y$ given random variables $(C,F)$, $H(Y\, | \,c_j,f_i)$ is entropy of $Y$ conditioned on specific values $(C=c_j,F=f_i)$ and $p(c_j,f_i)$ is the joint probability of a specific class/feature pair $(c_j,f_i)$. 

Now consider the flow of information through a set of one or more filters $F$ as quantified by the conditional entropy $H(Y\, | \,C,F)$. The success of the CNN lies in the backpropagation algorithm, which produces a set of filters $F$ that are highly discriminative regarding a set of training objects $C$. To be effective at discrimination, filters must be tuned to image structure characteristic of specific subsets of the objects to be classified. Intuitively, filters that identify the same structure equally well across all object classes are ultimately not useful in discriminating between object classes.

This intuition may be formalized for a filter set $F$ as follows. Let $\{C',C''\}$ represent a partition of set $C$ into mutually exclusive subsets of objects for which $F$ is informative and uninformative, respectively. The conditional entropy can be expressed as a binary sum over this partition:
\begin{align}
  H(Y\, | \,C,F) & \ = \ \\ \notag
  &p(C')H(Y\, | \,C',F) \ + \ p(C'') \, H(Y\, | \,C'',F),
\end{align}
where by definition:
\begin{align}
  H(Y\, | \,C',F) \ < \ H(Y\, | \,C'',F).
  \label{eq:entropy_difference}
\end{align}

In the case of classifying an image of a specific object $c_j$, the neural output $Y$ following a filter set $F$ is necessarily:
\begin{align}
   H(Y\, | \,c_j,F) \ = \
\begin{cases}
    \text{low},& \text{if } c_j \in C' \\
    \text{high},& \text{if } c_j \in C''.
\end{cases}
\end{align}
This analysis leads to the prediction that, for a set of highly discriminative filters $F$, e.g., resulting from CNN training via backpropagation, the conditional entropy $H(Y\, | \,C,F)$ of neural response $Y$ conditioned on $F$ serves as an excellent feature for discriminating between object classes. We refer to the conditional entropies $H(Y\, | \,C,F)$ computed across responses of filters or filter sets $F$ of a trained CNN as CENT features. In the following section, we demonstrate that they result in a powerful, compact code that can be computed at individual filters and/or filter layers throughout the network, and achieve better classification performance than the original CNN output itself.

\section{Experiments}

To emphasize the general usefulness of our CENT feature analsyis, we demonstrate CENT features in two distinct CNN classification contexts. The first is classifying Alzheimer's disease (AD) and age from volumetric MRI scans from the publicly available Open Access Series of Imaging Studies (OASIS) dataset~\cite{marcus2007open}. The second context is classifying natural object categories in 2D photographs, where experiments show that in the case of a pre-existing trained CNN, CENT features lead to markedly improved generalization performance in classify new, previously unseen categories, in comparison with the softmax output of the original CNN.

\subsection{3D Classification of Brain MRIs}

Analysis of neurodegeneration due to Alzheimer's disease or natural aging is a focus of significant interest, motivated by the need for developed nations to cope with an increasingly aging population demographic. In general, AD may be classified with high accuracy from brain MRI using precise measurements of cortical thickness~\cite{desikan2009automated}, however classification from whole-brain MRI data is still a challenging task~\cite{wachinger2016domain} where the maximum reported accuracy on the OASIS dataset is AUC=93.4\%~\cite{mahmood2013automatic}. A primary challenge is discriminating between older healthy subjects with natural age-related atrophy~\cite{toews2010feature} and younger diseased subjects with minimal atrophy. Entropy of derivative filters has been used as a feature in quantifying Alzheimer's in brain MRI~\cite{chaddad2016local}, however to our knowledge has not been applied to CNN classification. 

\begin{figure}[t]
\begin{center}
\includegraphics[width=0.6\linewidth]{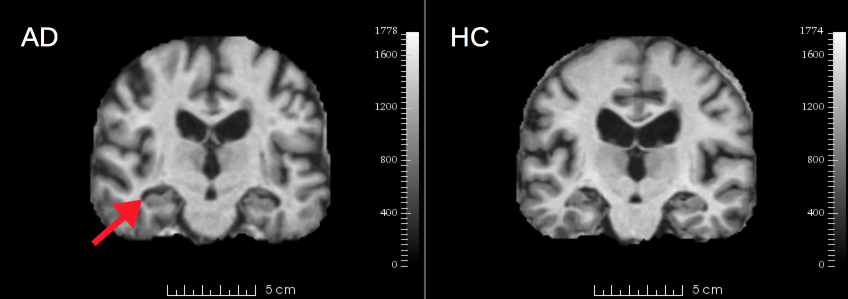}
\end{center}
   \caption{Coronal slices of 3D brain MRI illustrating a case of Alzheimer's disease (AD, left) and a healthy control subject (HC, right). A hallmark of AD is atrophy of the cortical surface (red arrow) surrounding the hippocampus, a neuroanatomical structure intimately linked short-term memory formation.}
\label{fig:ad_hc}
\end{figure}

Neuroimage classification here is based on the publicly available OASIS dataset~\cite{marcus2007open} consisting of brain MRIs from 416 unique, right-handed individuals spanning an age range of [18,96] years, with approximately equal numbers of male and females. We investigate binary classification in two contexts: 1) Alzheimer's disease (AD) vs. healthy controls (HC) 2) and old vs. young subjects. In general, both Alzheimer's disease and natural aging are characterized by cortical atrophy and enlargement of extra cerebral spaces. AD in particular is linked to cortical atrophy about the hippocampus, a neuroanatomical structure intimately linked to short-term memory storage which is impaired in AD sufferers. Figure~\ref{fig:ad_hc} illustrates subtle differences between AD and HC brains surrounding the hippocampus, and the challenge of AD vs. HC classification. 

\begin{figure*}[ht]
\begin{center}
\includegraphics[width=0.90\linewidth]{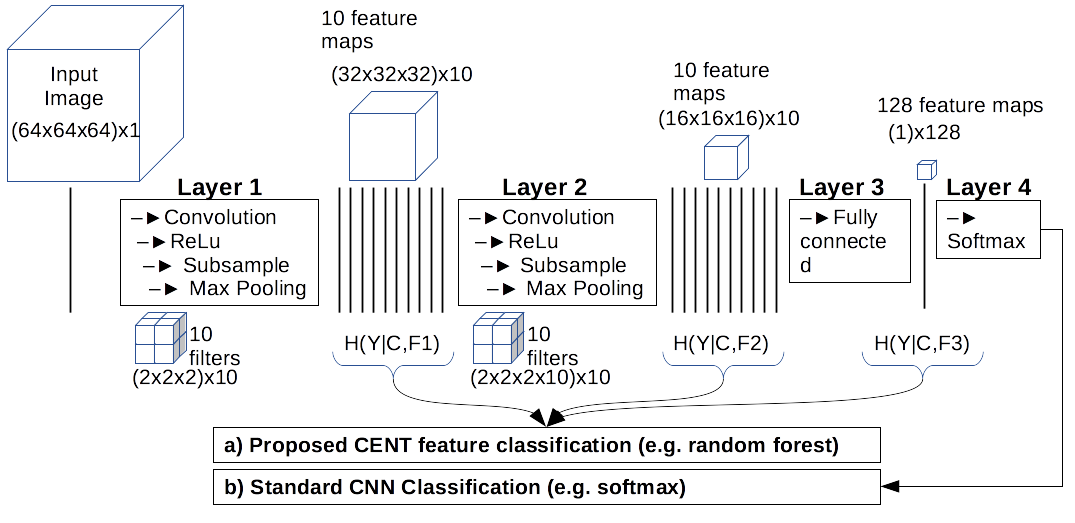}
\end{center}
   \caption{4-layer volumetric CNN architecture used in experiments, including 2 convolutional layers (with ReLU/subsampling/max pooling) followed by 1 fully connected layer and finally 1 softmax classification layer. In experiments, a) CENT features are computed from feature maps / output and passed to random forest classification, this is compared with b) standard CNN softmax output.}
\label{fig:network}
\end{figure*}

{\bf 3D CNN Architecture:} We developed a novel 4-layer CNN architecture based on 3D convolution filters as shown in Figure~\ref{fig:network}. The details are as follows. Input image size = $64 \times 64 \times 64$ voxels. Layer-1: filter size = $2 \times 2 \times 2$; stride=2; filters=10; Max pooling; ReLU; output = 10 feature maps of size ($32\times 32 \times 32$). Layer-2: filter size = $2 \times 2 \times 2$; stride=2; filters=10; Max pooling; ReLU; output = 10 feature maps of size ($16\times 16 \times 16$). Layer-3: fully connected layer; output = vector size 128. Layer-4: softmax = vector size 2; \footnote{The MDCNN Matlab CNN implementation is used https://www.mathworks.com/matlabcentral/fileexchange/58447-hagaygarty-mdcnn}.

We compute conditional Shannon entropy from the first 3 CNN layers by aggregating filter responses into 256-bin histograms using the Shannon entropy. Conditional entropy features for classification are generated from individual feature map responses (10+10+1=21 CENT features) and from responses aggregated at each layer (1+1+1=3 CENT features). 

{\bf Alzheimer's Disease vs. Healthy Brains:} The goal of experiments here is to classify between AD and HC subjects from MRI data. The relevant data subset used consists of a total of 198 (AD=100, HC=98) MRI scans from demographically matched subjects with age $\geq$ 60 years. The 3D CNN is trained from 50 AD and 50 HC subject MR images and classification is then tested using the remaining subjects in a 5-fold cross validation. Softmax classification is performed in a standard fashion from the CNN output. CENT feature classification is carried out using a random forest (RF) classifier with 100 trees. We chose to use the RF as it is one of the most effective and general-purpose classification algorithms, running efficiently on large databases with thousands of input variable/features~\cite{breiman2001random}, other classifiers could be used. We then performed the 5-fold cross-validation strategy to obtained unbiased estimates of performance, where training images (i.e. features) are divided into 5 equal sized subsets and, in each fold, one subset is put aside for testing and the remaining 4 subsets are used to train the RF classifier. Finally, the AUC value is computed as the average AUC obtained across all 5 folds.

\begin{figure}[t]
\begin{center}
   \includegraphics[width=0.6\linewidth]{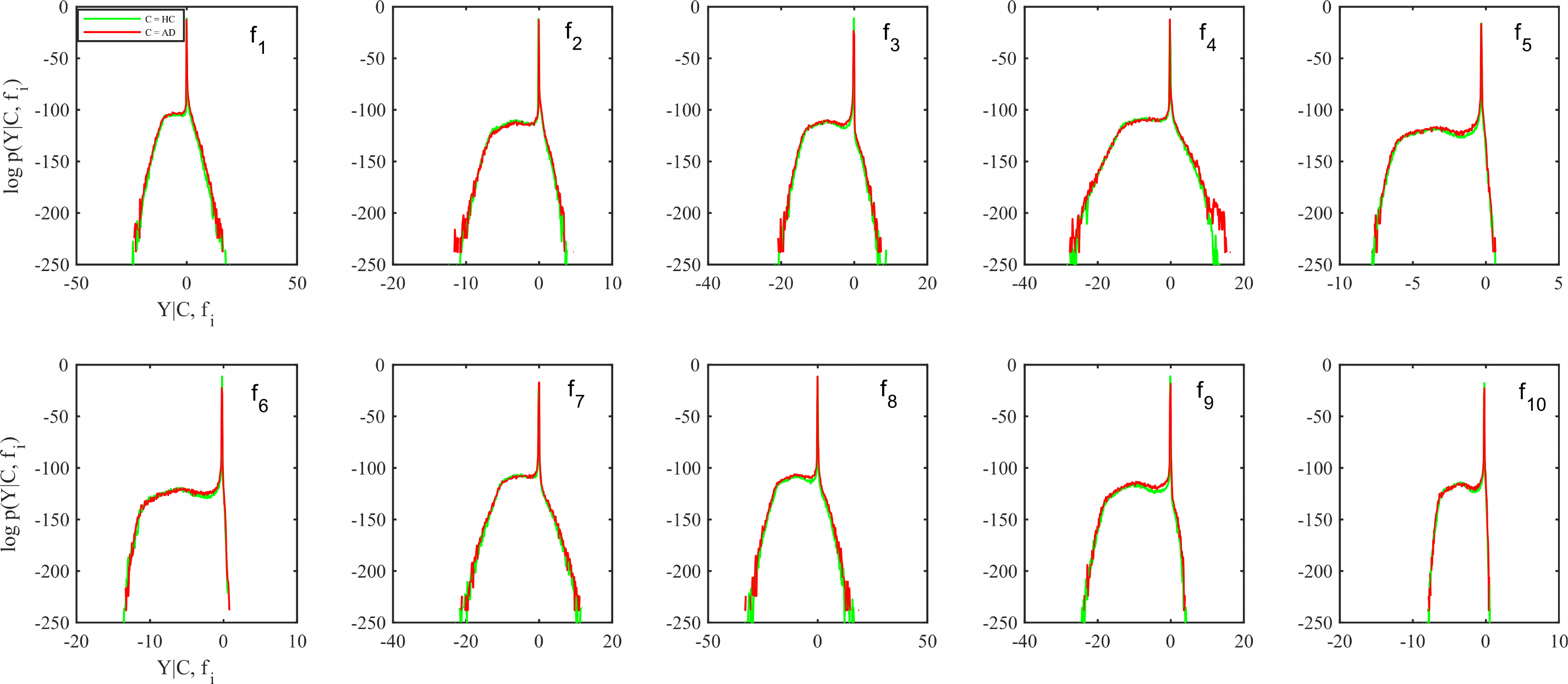}
\end{center}
   \caption{Response distributions $p(Y\, | \,C,f_i)$ for 10 convolutional features maps in layer 2, note that the vertical axis is displayed in logarithmic units.}
\label{fig:neuron_response}
\end{figure}

Figure~\ref{fig:neuron_response} shows the distributions of convolution responses from which CENT features are computed. All distributions show heavy-tailed characteristics centered around zero, reminiscent of filters derived from natural images~\cite{simoncelli2001natural,simon2007scene}.

\begin{figure}[t]
\begin{center}
\begin{tabular}{cc}
\includegraphics[width=0.3\linewidth]{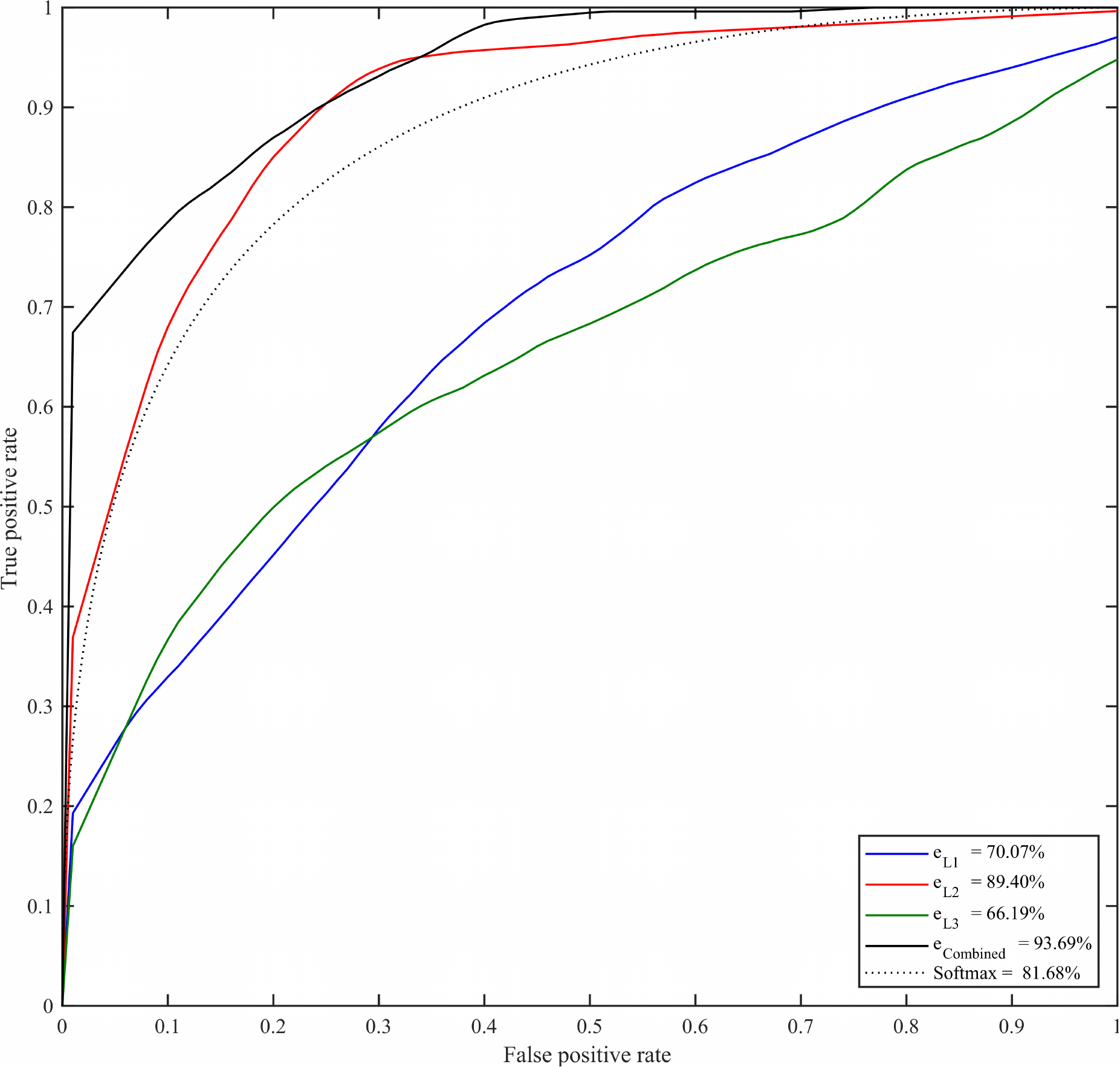} &
\includegraphics[width=0.4\linewidth]{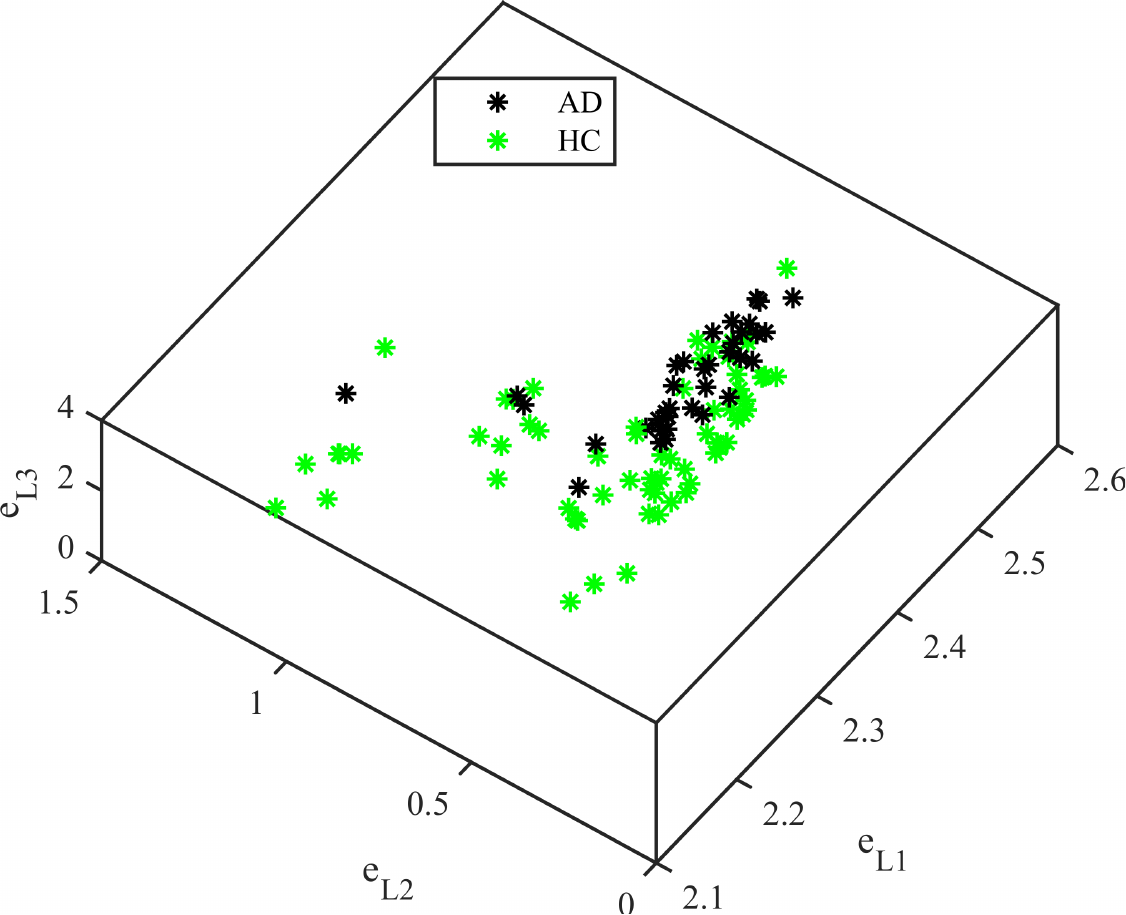} \\
(a) & (b)
\end{tabular}
\end{center}
   \caption{(a) ROC curve for AD vs. HC classification using layer-wise CENT features. (b) Scatter plot of 3 CENT features showing clear separation between AD and HC classes, the 3 axes represent conditional entropy computed at the 3 CNN layers.}
\label{fig:ad_cent_layerwise}
\end{figure}

Figure~\ref{fig:ad_cent_layerwise} (a) shows the ROC curve for 3 CENT features computed at CNN layers, both individually and combined. The highest overall AUC=93.6\% is obtained from a classifier combining the 3 CENT features. This is the highest reported classification result in the literature for AD classification from the OASIS brain MRI database. Surprisingly, it is 12\% higher that the output of the softmax classification  of the original trained CNN. Figure~\ref{fig:ad_cent_layerwise} (b) shows a scatter plot of the 3 CENT feature values for subject categories, where the class separation is clearly visible.

\begin{figure}[t]
\begin{center}
\includegraphics[width=0.35\linewidth]{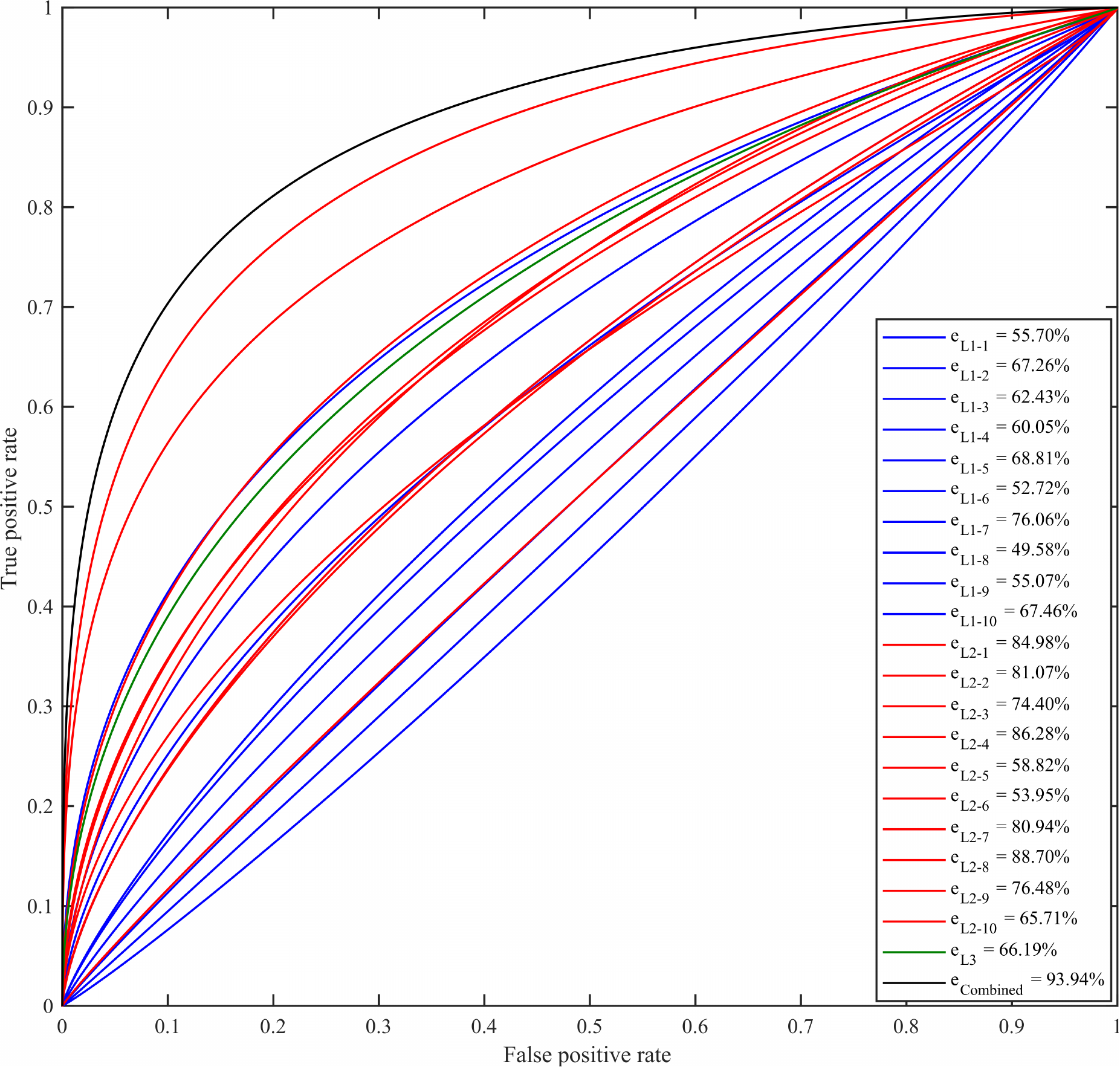}
\end{center}
   \caption{ROC curve for AD vs. HC classification using filter-wise CENT features, with 10 filters in layer 1 (blue curves), 10 filters in layer 2 (red curves) and 1 filter in layer 3 (green curve). As in layer-wise CENT classification of Alzheimer's, the most informative features are generally found in layer 2, and highest classification is obtained by combined all features (black curve).}
\label{fig:ad_cent_featurewise}
\end{figure}

Figure~\ref{fig:ad_cent_featurewise} shows the result of classification using CENT features computed on a filter-wise basis. Individually, features computed from the 10 filters in layer 2 (red curves) result in generally higher classification than the 10 in layer one (blue), the single fully connected layer 3 (green) results in intermediate classification performance. Again, combined classification based on all 21 features (black) results in the highest AUC=93.94\%, a similar value obtained using layer-wise features.

\begin{figure}[t]
\begin{center}
\includegraphics[width=0.25\linewidth]{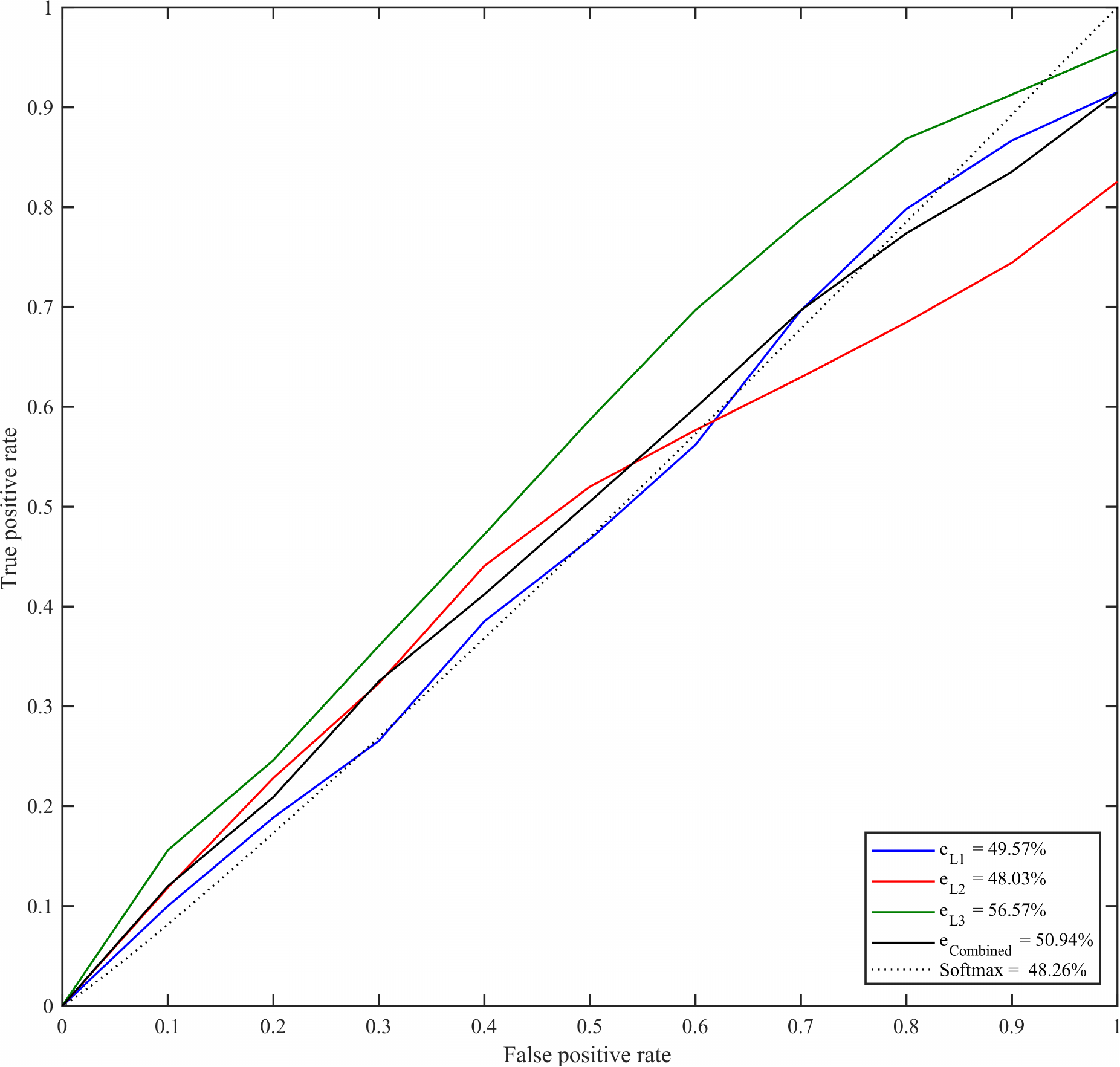}
\end{center}
   \caption{ROC curve for AD vs. HC classification following randomly permuted class labels and 3 layer-wise CENT features.}
\label{fig:permutation_cent_layerwise}
\end{figure}

{\bf Random Alzheimer's Label Permutations:} As a baseline, we test AD-HC classification following label permutations, where we expect to observe uninformative classification. The CNN is trained and tested as before. Figure~\ref{fig:permutation_cent_layerwise} shows the result of classification, which is uninformative as expected.

{\bf Young vs. Old Brains:} The goal here is to test MRI brain classification on a variable other than disease, young and old age groups. We considered the set of 329 MRIs from healthy subjects, partitioned into young and old classes using the median age. 200 images were used to train the CNN (100 old and 100 young) and classification is evaluated on the remaining 129 subjects (68 old, 61 young), again using 5-fold cross validation with the random forest classifier.

\begin{figure}[t]
\begin{center}
\begin{tabular}{cc}
\includegraphics[width=0.3\linewidth]{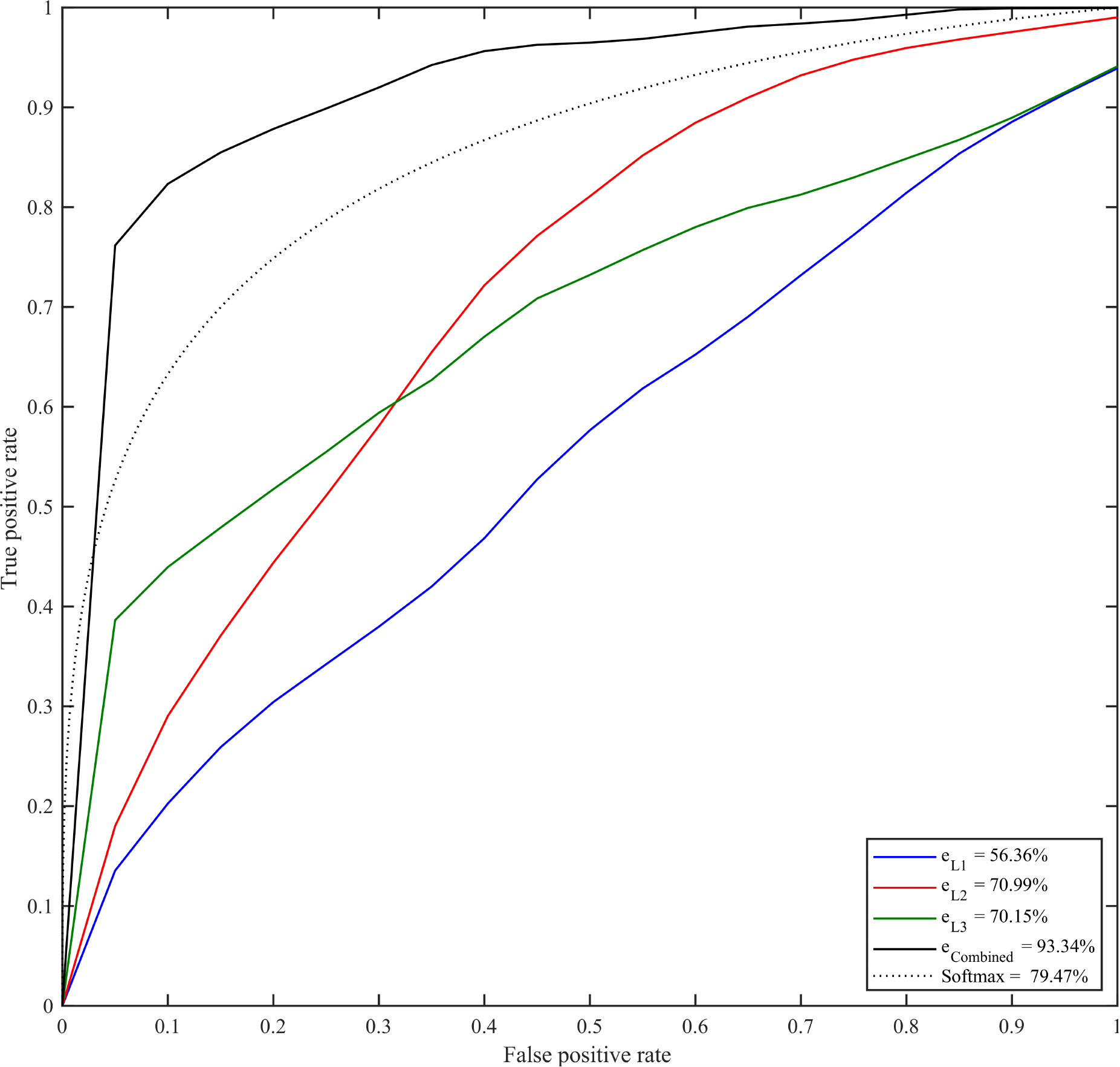} &
\vspace{0.5cm}
\includegraphics[width=0.4\linewidth]{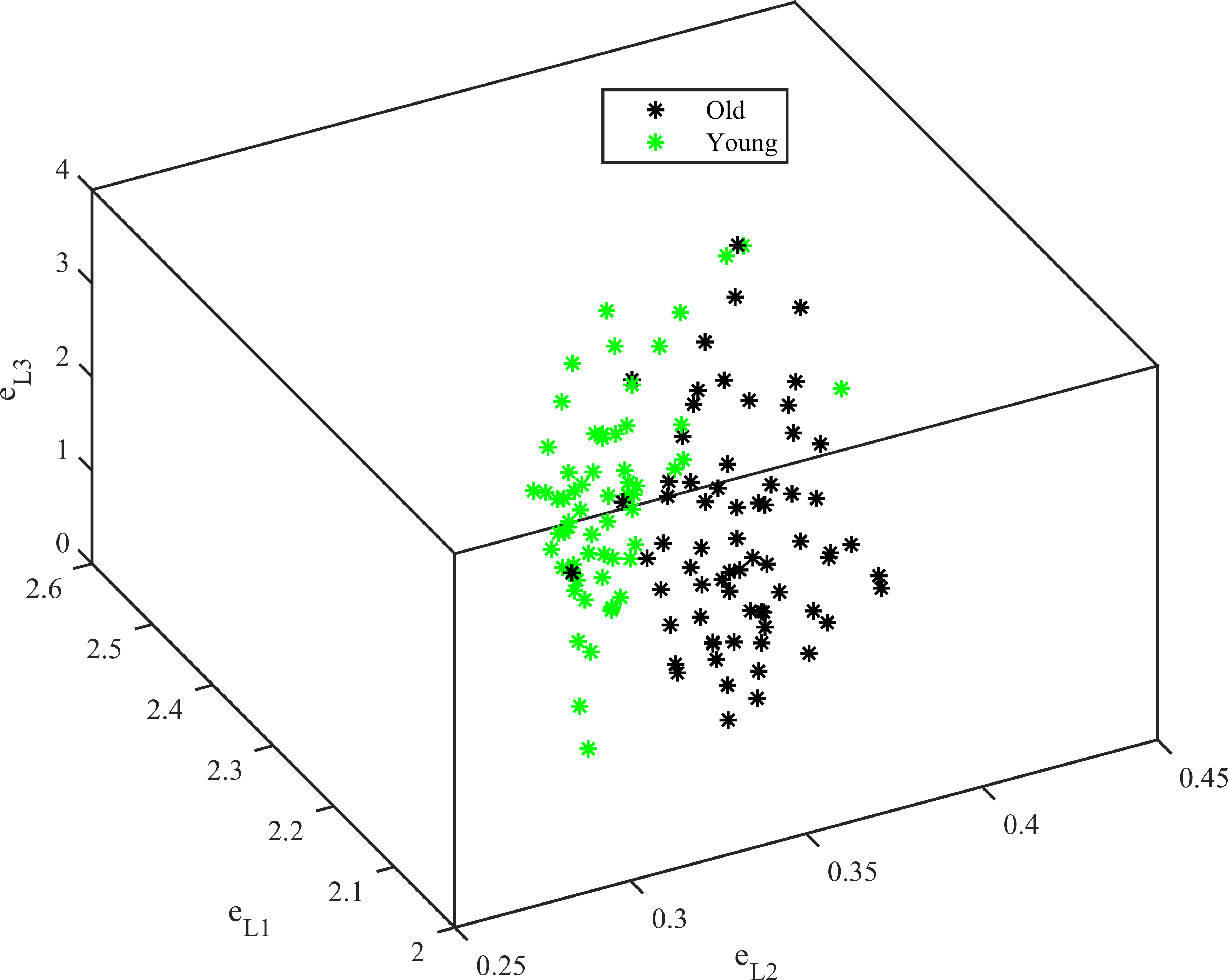} \\
(a) & (b) \\
\end{tabular}
\end{center}
   \caption{(a) ROC curve for CENT feature classification of young vs. old subjects. (b) 3D plot of per-layer conditional entropy showing clear class separation.}
\label{fig:age_cent_layerwise}
\end{figure}

Figure~\ref{fig:age_cent_layerwise} shows the result of age classification based on 3 layer-wise CENT features. Here, the fully connected 3rd layer results in the best individual performance (green). The combined performance of all 3 CENT features is highest at AUC=93.34\% (black), almost 14\% higher than the softmax output of the CNN.

\subsection{2D Classification of Visual Object Classes} 

As a final experiment, we demonstrate CENT features in the context of transfer learning for natural object categories, where the goal is to adapt the responses of an existing CNN to classify images arising from previously unseen object categories. For this, we consider the existing VGG CNN trained with the 1000 object categories~\cite{chatfield2014return} from the ImageNet dataset~\cite{deng2009imagenet}\footnote{The MatConvNet Matlab software package is used http://www.vlfeat.org/matconvnet/}, in general any CNN trained on a large set of diverse image data would suffice. To test classification via transfer learning, we select an arbitrary set of 10 object categories (30 images each) from the Caltech 101 data~\cite{fei2007learning} that are not found in the ImageNet data used in CNN training. These are 1) anchor, 2) buddha, 3) chandelier, 4) snoopy, 5) gramophone, 6) lotus 7) metronome, 8) minaret, 9) stapler and 10) yin-yang.

Each test image is passed through the CNN, and two types of features are computed for investigating transfer learning: 1) CENT features and 2) standard CNN softmax output. 1) CENT features are computed across filter outputs at convolutional network layers. Using the 'matconvnet-vgg-f' CNN~\cite{chatfield2014return} we compute CENT features from filters in the first 5 convolutional layers, for a total of 64+4*256=1088 CENT features. 2) Standard CNN softmax features are taken as the 1000-element vector over the 1000 training classes. Other CNN layers could potentially also be used in transfer learning, here the CNN output should represent a suitably informative code over previously unseen classes.

\begin{figure}[t]
\begin{center}
\begin{tabular}{cc}
\includegraphics[width=0.40\linewidth]{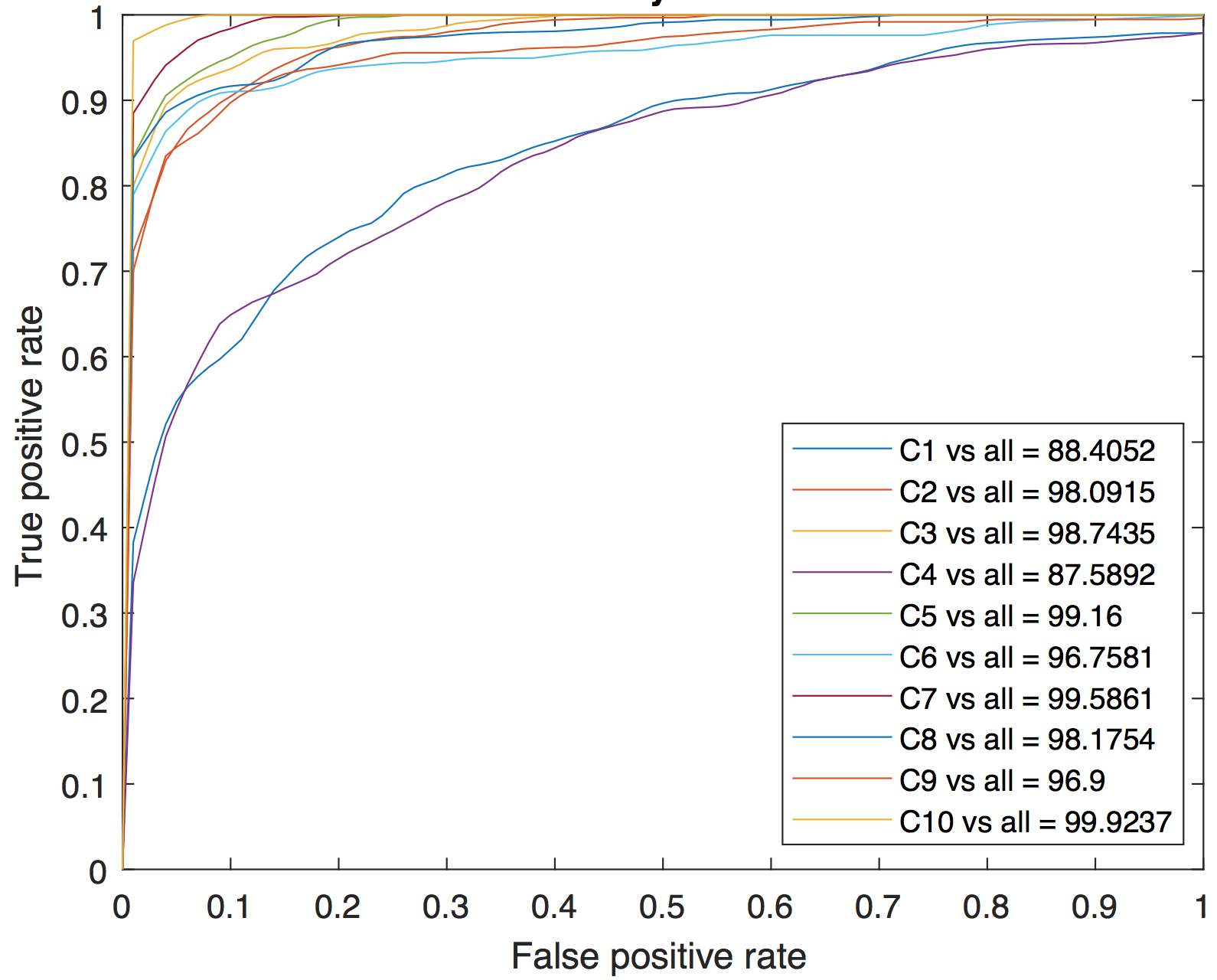} & 
\includegraphics[width=0.40\linewidth]{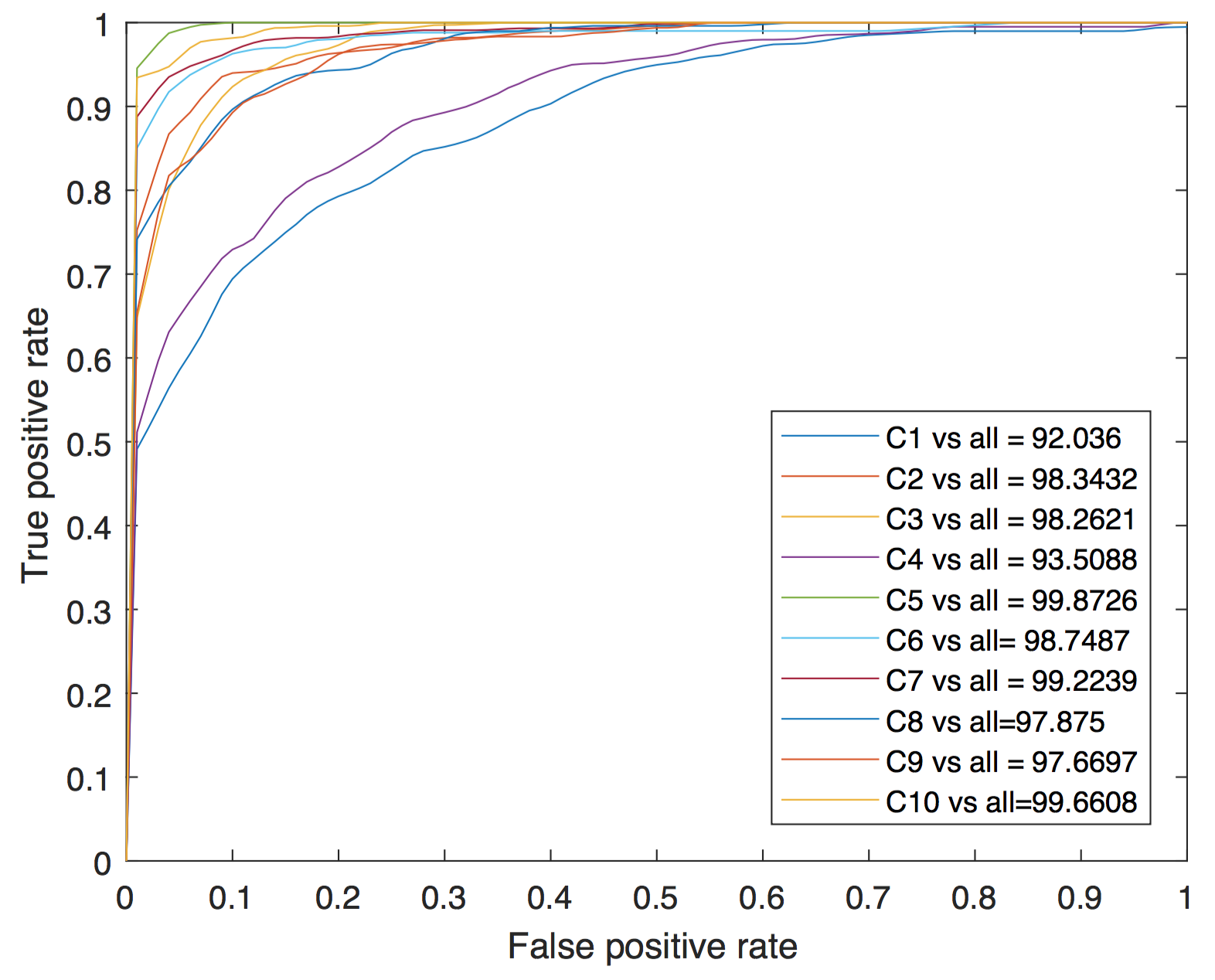} \\
(a) ROC curve for CENT features &
(b) ROC curve for CNN softmax features
\end{tabular}
\end{center}
   \caption{ROC curves transfer learning: classification of 10 natural objects not used in original CNN training. Each image is represented by (a) CENT features computed layer-wise across 5 CNN convolutional layers and (b) the 1000-feature CNN softmax output.}
\label{fig:objects_cent_vs_cnn_softmax}
\end{figure}

Random forest classification with 5-fold cross validation is used to evaluate one-vs-all classification for a) CENT features and b) standard CNN softmax output. Default RF settings are used, where the maximum number of splitting variables is equal to the square root of the number of total variables, here $32 \approx \sqrt{1000}$ for both CENT and softmax. Figure~\ref{fig:objects_cent_vs_cnn_softmax} shows ROC curves obtained by classification. We observe virtually identical AUC values for both configurations. This demonstrates the high degree of information concentrated in CENT features, and suggests that an effective means of modeling previously unseen classes may be via the conditional entropy of filter outputs throughout the CNN, rather than the CNN output layer itself. The CENT features selected as the basis for RF decision tree splits correspond the CNN filers most information regarding the class of interest. These are a unique combination of CNN filters unique to each object category, and come primarily from deeper layers in the network.

\section{Discussion}

This paper provides a principled analysis of information flow through convolutional neural networks using information theory. We show that a discriminatively trained network filter set $F$ leads to a reduction in conditional entropy $H(Y\, | \,C,F)$, that is necessarily greater for the subset of object classes for which the filter set is tuned. Both theory and experiments show that the conditional entropy of filter responses, CENT features, result in a highly compact, informative code for image classification.

In experiments using brain MRI data, 3 CENT features computed from each CNN layer result in the highest whole-brain classification rate of Alzheimer's disease reported for the OASIS dataset, with an AUC=93.6\%. Most surprisingly, this is 12\% higher than the fully connected softmax output of the the original CNN trained for the task. This success appears to be a consequence of integrating information throughout the network into classification, rather than in a sequential fashion at the output, as predicted by the data processing inequality~\cite{cover2012elements}. While this is similar in spirit to the densely connected network approach~\cite{huang2016densely}, it is achieved here with only 3 CENT features, rather than thousands of additional dense connections. A similarly high AUC value is achieved for classifying brain MRIs into young and old age categories. 

Experiments in transfer learning from 2D photographs show that CENT features computed filters in an existing trained CNN can be used to achieve effective classification for 10 object categories not found in the training data, with ROC AUC values similar to the 1000-element softmax layer at the output of the original CNN. This performance indicates a promising avenue for low-parameter transfer learning.

It is remarkable that a relatively small number of CENT features computed throughout the network can lead to effective classification. Intuitively, it appears that each CENT feature serves as a bit of information to constrain the identity of the object class, as predicted by a class-discriminative filter set. Thus in the case of ideally discriminative filters capable of partitioning an image unambiguously into one of $N$ categories, a minimum of $\log_2 N$ filters would be required to provide a unique code to each category. This may provide insight into the expected growth of CNN size with respect to the number of object categories, and is an avenue of future investigation.

The idea of analyzing information flow through a neural network using information theory is not new, however to our knowledge it has not been rigorously applied in recent work with deep CNNs. Empirically, one of the reasons why it works so well with modern CNN architectures appears to be that convolution outputs $Y$ are highly normalized/tuned, and may be aggregated across features maps and layers into reliable features. We note that computation of CENT features after vs. before ReLU normalization has a significant improvement on classification. Without ReLU, there is the possibility that bimodal responses $Y$ may be produced for different classes, i.e. highly positive and negative correlations for two different groups of object classes, thus weakening the argument of distinctive class-informative filters $F$.

Future work will involve further investigation of CENT features in the context of natural object classification from large scale datasets. Conditional entropy computed throughout the network may lead to new optimization techniques in CNN training which seek to maximize information gain of individual filters of filtering layers. Likewise, conditional entropy computed in a spatially localized fashion may prove useful in identifying image regions most informative regarding classification.

\end{document}